*Review*
**From Biological Synapses to "Intelligent" Robots**

**Birgitta Dresp-Langley**
CNRS UMR 7357 ICube Research Department, Strasbourg University, 67081 Strasbourg, France; birgitta.dresp@cnrs.fr

**Abstract:** This selective review explores biologically inspired learning as a model for intelligent robot control and sensing technology on the basis of specific examples. Hebbian synaptic learning is discussed as a functionally relevant model for machine learning and intelligence, as explained on the basis of examples from the highly plastic biological neural networks of invertebrates and vertebrates. Its potential for adaptive learning and control without supervision, the generation of functional complexity, and control architectures based on self-organization is brought forward. Learning without prior knowledge based on excitatory and inhibitory neural mechanisms accounts for the process through which survival-relevant or task-relevant representations are either reinforced or suppressed. The basic mechanisms of unsupervised biological learning drive synaptic plasticity and adaptation for behavioral success in living brains with different levels of complexity. The insights collected here point toward the Hebbian model as a choice solution for "intelligent" robotics and sensor systems.

## 1. Introduction

The Hebbian synapse and synaptic learning rules [1] are the fundamental conceptual basis of unsupervised learning in biological and artificial neural networks [2]. A synapse refers to a connection between two neurons in a biological or artificial neural network, where the neuron transmitting information via a synapse or synaptic connection is referred as the presynaptic neuron, and the neuron receiving the information at the other end of a synaptic connection as the postsynaptic neuron. The information propagation, and its efficiency, of biological and artificial synapses is strictly self-reinforcing, following a principle called self-organization, which is explained in further detail in Section 4.3. The more a synapse is stimulated, the more effectively information flows through the connection, which ultimately results in what Hebb [1] and subsequently others have called the long-term potentiation (LTP) of neural connections. Synaptic connections that are no longer repeatedly stimulated and, as a consequence, no longer self-reinforced will lose their information propagation efficiency, which ultimately results in the long-term depression (LTD) of neural connections. A schematic illustration of synaptic learning is shown below (Figure 1). This selective review starts with a brief recall of the principles of Hebbian synapse-based learning (Section 2). On this basis, specific examples of biological learning in vertebrates and invertebrates (Sections 3 and 4) are then brought to the forefront to illustrate the potential for bioinspired neural network models and self-organizing control of simple and complex agentic functions of robots or other artificially intelligent system. Such functions include rhythmic movement generation and control, goal-directed behaviors, task space coding, sequential action timing, alternative event choice, and sensorimotor integration for action. Such functions are then discussed using examples from current developments in robotics (Section 5) to further clarify how converging sensory and reinforcement (reward) learning can make a functional network as a whole capable of acquiring task structures and self-organizing toward further learning. The conclusions include a summary table (Table 1) with references to specific models (with equations) of self-organization as a function of the type of behavior (function), model organism (species), and level of complexity for "intelligent" robot design architectures.

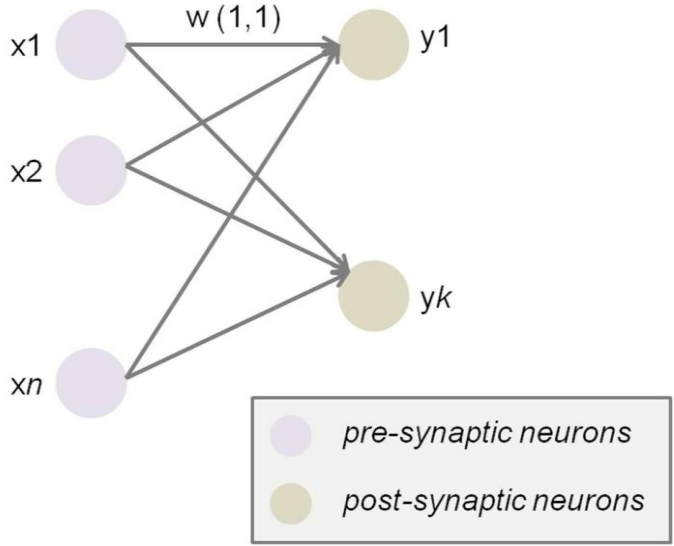

**Figure 1.** Schematic illustration of synapses within a small neural network. Self-reinforcing (Hebbian) synaptic learning leads to the progressive increment of the synaptic weights (w) of efficiently stimulated neural connections, which are thereby long-term potentiated. The connectivity of nonreinforced synapses weakens and, ultimately, becomes long-term depressed. The synaptic learning rules are the fundamental conceptual basis of unsupervised reinforcement learning in biological and artificial intelligence.

**Table 1.** Levels of robotic agency and species serving as model (I = invertebrate, V = vertebrate) for self-organized control architectures in increasing order of functional complexity (NN = neural network).

| Functional Complexity | Agency | Species | Control Level | Implemented | Selected References |
|---|---|---|---|---|---|
| Single NN | Rhythmic movement | I | Self-organized | Yes | [42,44,46,54] |
| Single NN | Goal-directed action | I, V | Self-organized | Yes | [30,64,68,70–72,80–83,91,98] |
| Single NN | Alternative choice | I, V | Self-organized | Yes | [10,25,53,62] |
| Single NN | Sequenced action | I, V | Self-organized | Yes | [46,98,99,105,128,150,152–155] |
| Multiple NN | Sensorimotor integration | I, V | Self-organized | Yes | [53,62,114,115,128,129131,132] |
| Multiple NN | Cognitive planning | V | Self-organized | Partially | [2,3,98,99,150] |
| Multiple NN | Voluntary action | V | Self-organized | No | [2,3,98,151] |

## 2. Biological Synapse: Adaptive Learning "from Scratch"

Any object of the physical world is defined by multiple features and properties such as shape, texture, luminance, color, weight, taste, sound, or function. Each such feature is represented via different modalities in interconnected cortical regions of the mammalian brain [2–4]. The brains of living systems learn about such physical regularities "from scratch", i.e., without any prior knowledge, through unsupervised mechanisms. The Hebbian learning principles generate distributed multimodal brain representations, as clarified further below, in networks of functionally connected neurons, each contributing to specific sensory and/or motor processes related to an object and generating short-range representations in dedicated neural networks (populations) that interact with each other at long-range spatial scales.

*2.1. From Single Synapses to Brain Networks*

According to Hebb [1], each type of cell assembly has a functionally specific connectivity, thereby acquiring the status of a functionally dedicated neural network selective for a particular sensorial or cognitive function, or a particular type of information. The propagation of information during synaptic learning may be event-driven [5], clock-driven [6], or a combination of both [7,8]. The general basis of all computational development in this regard is a simplified synapse model, where the spike input will trigger a synaptic electric current into the postsynaptic neuron. The Hebbian learning principle in its most general form is expressed in mathematical terms as

$$\Delta w_{ij} \propto v_i v_j, \tag{1}$$

where $w_{ij}$ refers to the change of synaptic weight between the presynaptic neuron $i$ and the postsynaptic cell $j$, and $v$ represents the activities of those neurons, respectively. Any network of strongly connected neurons in a functionally dedicated neural network or cell assembly may communicate with another functionally dedicated network to generate multimodal brain representation. According to Hebb [1], the combined activity of functionally specific networks in the mammalian brain explains the full complexity of cognitive representation ("intelligence", "intelligent processes") on the basis of a from-simple-to-complex processing hierarchy. Functionally connected neurons act as a functional unit, with the activation of a fraction of that unit leading to the activation of the whole unit. When no longer activated, the weight of synaptic connections is weakened and, ultimately, the functional connectivity extinguishes as a result of the mechanisms of LTP and LTD that govern biological neural learning. LTP and LTD are triggered by the timing of neural signals (spikes) in the short-range spatial regime governing interactions between adjacent neurons. The hypothesis that the same timing principles apply to the long-range regime of functional interaction between neurons across distant cortical areas is supported by functional neuroanatomy and psychophysics [3,4].

*2.2. Timing of Neural Signals*

The timing of neural signals in a network [5–8] determines whether neural connections are reinforced (excitation) or suppressed (inhibition). When a presynaptic signal precedes a postsynaptic signal, potentiation of the synapse resulting in a stronger weight $w$ is observed, with repeated strengthening ultimately leading to long-term potentiation (LTP); a repeated reverse temporal signal order weakens synaptic strength and, ultimately, leads to long-term depression (LTD). Because of the absence of explicit goals, correction functions, or prior knowledge, Hebbian synaptic learning is categorized as unsupervised learning. The information propagation in such networks may be event-driven [5], clock-driven [6], or a combination of both [7,8]. In self-organizing reinforcement learning, the weight $w$ of an eligible synapse $c$ changes in time $t$ with the reinforcement signal $R$.

$$w(t) = R(t) \times c(t), \tag{2}$$

where *R(t)* is the "reward" signal at a given moment in time *t*. Starting from these basic functional principles, LTP and LTD promote increasingly effective functional organization in neural networks such as those found in biological brains. Neural encoding therein is to represent information from the physical world, such as the direction of object or limb movement, in the activity of a neuron (spike activity, firing rate). Information decoding is the reverse process of interpretation of neuronal activity and its translation into electrical signal for actuators (such as muscles, digits, or limbs). The biological brain encodes information in two continua corresponding to physical space and neural space. The physical space may be the physical properties of objects such as their color, shape, or temperature; the neural space consists of functional properties of a neuron such as firing rates and peaks. In binary coding, neuron models take two values corresponding to on/off states while ignoring the timing and multiplicity of spikes. Binary coding [9] converts the spike timing of neurons during specific time intervals into binary codes and is, for example, used to represent categories during learning [10]. Rate coding, on the other hand, is based on the coding of intensity of sensory stimuli [11]. Often a population or assembly of neurons is used as a whole to encode specific information, a strategy that is consistent with that of brains in living agents, where specific functions are controlled by a specific class of neurons.

*2.3. Reinforcement and Extinction*

One of the ambitions of biologically inspired neural coding in models for robot control is to characterize and provide a plausible brain model for the behavioral neuroscience of reinforcement and extinction [12,13].

These are phenomena in which a behavior that has been acquired through reinforcement in operant learning decreases in strength until its full extinction when the outcome or event that reinforced it is no longer occurring (Figure 2). The mechanisms of reinforcement and extinction appear to involve three functionally identified regions of neural circuitry formed by the *amygdala*, the prefrontal cortex, and the hippocampus [14]. The remainder of this article discusses the principles of reinforcement and extinction in biological learning using examples. These were selected on the basis of the self-organizing mechanisms brought to the fore in the work cited. How their implementation in artificial neural network architectures can promote the design of intelligent robots and sensor technology is then made clear. An essential ground condition of self-organizing intelligence is the functional plasticity of neural networks.

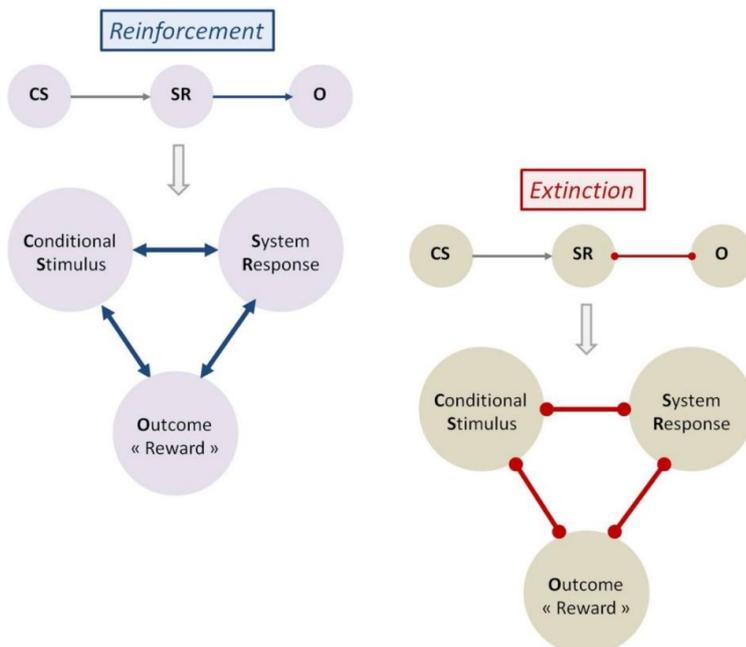

**Figure 2.** Phenomena of reinforcement (**left**) and extinction (**right**) account for all learning in brain and behavior. A system response (SR) to a conditional stimulus (CS) steadily reinforced by a specific outcome (O) or "reward" during learning leads to a consolidated functional network connectivity (**left**). Such connectivity decreases in strength until its full extinction when the outcome that initially reinforced it is no longer delivered (**right**).

*2.4. Functional Plasticity*

Throughout the process of synaptic learning and memory consolidation, structural changes driven by activation of one or several neurotransmitter receptors take place in the target neural networks [15]. These neurobiological changes are the basis of all functional plasticity. The most important excitatory neurotransmitter system in this respect would probably be the glutamatergic system [16], since its involvement in persistent forms of synaptic plasticity is well recognized [17]. After the activation of neurotransmitter receptors, several downstream signals are triggered. The most important signal for synaptic learning is calcium, which has the ability to interact with the actin cytoskeletons of dendrites and, through this interaction, regulates structural and, as a consequence, functional synaptic plasticity [18]. After synaptic activation, the flow of calcium ions ($Ca^{2+}$) into cells, either through gated calcium channels or via internal reservoirs, results in complex series of transitory oscillatory signals [19]. Such signaling complexity needs to be transformed into stable and persistent messages, which explains the need for self-organizing structural change in biological neural networks [20]. In a relatively constant environment, animals may express variable behaviors or motor actions as a consequence of internal drives and motivations. Such actions are driven by adaptive pressure, i.e., the need to survive in a changing environment, and arise from the dynamic properties of so-called central networks, i.e., the control structures of a given function, behavior, or agency, in a living brain. Such adaptive processes may result in stereotyped behaviors and action patterns or in highly variable and goal-directed choice behaviors [21–24,26]. Feeding, sexual, and aggressive behaviors in invertebrates and vertebrates are goal-directed actions relying on lesser or higher degrees of functional complexity in which internal "decisions" to act determine the spontaneous expression of survival relevant actions and activity patterns [26–29]. "Decision to act" here implies that the neural network has structural and functional mechanisms which enable the selection of a particular behavior, action, or activity pattern from several variants thereof. Such mechanisms internally represent the conditions for external expression in the form of action. The network mechanisms that govern this kind of internal decision making are subject to plastic changes by self-organization, i.e., through regulation by changes in sensory inputs from the outside world associated with the positive (reward) or negative (punishment) consequences of a specific action, behavior, or response [30–32]. Sensory feedback in associative learning enables memory representation and modification of internal motivation [21–32]. With increasing structural and functional complexity of neural network connectivity, internal feedback mechanisms become increasingly essential for determining the relevance ("value", "meaning") of an external context for the production of a specific response or activity pattern [33–35]. Research data pointing toward specific neural structures implicated in such internal, behaviorally relevant decision making in living organisms are available [36–39]. However, the functional principles of self-organization that account for the capability of a neural network or a set of multiple inter-connected networks to generate and organize neuronal activity for coherent action-in-context selection, spontaneously and at different levels of complexity, have remained the holy grail in functional and computational neuroscience. A related and still unresolved question is how internal decision-making processes are regulated by further learning and long-term memory changes.

## 3. Invertebrate Models of Adaptive Learning

Invertebrate mechanisms of learning and memory illustrate that the biological brain mechanisms that control learning have a long evolutionary history. Living beings may be conceived as evolving, developing agents with a need to cope with environmental uncertainty. Most of the current knowledge of the central nervous system (brain) to spontaneously trigger motor behavior and actions that follow a specific pattern stems from the analysis of rhythmic, largely stereotyped, behavior of invertebrates. Limb and body movements (locomotion) and respiratory activities (breathing patterns) are examples of such stereotypes. So-called "rhythmogenic networks" or "central pattern generators" have been proposed to account for the synaptic and intrinsic membrane properties of neurons governing stereotypic behaviors [40–44]. Ongoing operations can be dynamically regulated or modulated in such networks by sensory input, but their functional variability is not necessarily determined by internally represented motivational components [45,46]. Invertebrate neural network models provide a mechanistic account for this kind of "low-level" learning, with and/or without internal representation [26–31]. Insights into the functional design and properties of motor networks able to autonomously elaborate action patterns and context setting for their expression [47] mainly stem from the neuroscience of invertebrate organisms [24–30], suggesting model circuits for low-level command processes governing invertebrate behavior(s). Experimental and model data have allowed characterizing the synaptic organization, cellular properties, and dynamic network organization in invertebrates including mollusks [48–50].

### 3.1. Motor Learning and Memory

The sea snail *Aplysia* is among the most widely used 'model organisms' in the cellular biology of low-level motor learning and memory. For his contributions to this field of discovery, Eric Kandel [51] shared the 2000 Nobel prize for Physiology or Medicine. *Aplysia* has two functionally identified motor neurons (Figure 3); the large size of these neurons correlates with the sensory areas they connect to, and each neuron can act as a single integrative center for the control of multiple motor behaviors. This species, thus, exploits a distinct strategy from others where complex tasks are controlled by several thousands of neurons. *Aplysia* can develop both non-associative and associative forms of long-term memory [48] needed for all fundamental learning (habituation, sensitization, classical and operant conditioning). The cellular and molecular mechanisms of long-term plasticity in *Aplysia* have many parallels in humans, which suggests a profound evolutionary conservation of the most elementary events underlying all learning and memory [51]. This has important implications for biologically inspired artificial intelligence for robot movement control, as made clear later. Until recently, learning and memory in invertebrate organisms were believed to be mediated by relatively simple presynaptic mechanisms. New experimental evidence from research using *Aplysia* indicates that the previously defended distinction between invertebrate and vertebrate synaptic mechanisms of learning is invalid. Learning in *Aplysia* cannot be explained in terms of presynaptic mechanisms only, given that NMDA receptor-dependent LTP appears to be necessary for classical conditioning in *Aplysia* [30]. Moreover, modulation of postsynaptic ionotropic glutamate receptor trafficking underlies behavioral sensitization in this snail [52]. Exclusively presynaptic processes drive relatively brief memory in *Aplysia*; more persistent memory forms in invertebrates are, therefore, likely to be mediated by postsynaptic processes as in vertebrates or by presynaptic mechanisms that depend on feedback signals [52,53], which results in the same outcome. In short, the neuronal underpinnings of variable motor strategies employed by simple living organisms such as invertebrates already depend, at least partly, on autonomous neural mechanisms, i.e., self-organization. The structural and functional properties of the networks mediating invertebrate motor activity spontaneously select external conditions for the expression of distinct, sometimes opposing, actions [48,54–57]. Invertebrate neural networks are capable of driving extinction learning [52–58], which relates to the ability to update previously learned information by integrating novel and, in essence, contradictory information. Such relearning has an important adaptive function and relevance for artificial intelligence design approaches in robotics.

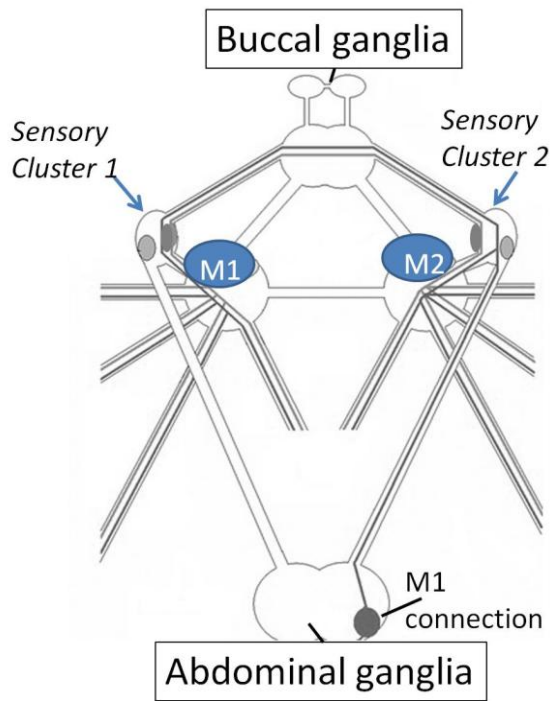

**Figure 3.** Snapshot view of the neuroanatomy of *Aplysia* central nervous system. This invertebrate exploits a distinctly different coding strategy compared with higher-order species, where behavior is controlled by several thousands of neurons. In *Aplysia*, synaptic actions attributed to neural activities governing positive and negative responses in feeding behavior are produced by large spontaneously active motor neurons (M), of which two have been functionally mapped in the cerebral ganglia. These motor neurons have an unusually large soma and act as multi-action interneurons, generating inhibitory and/or excitatory synaptic potentials in connected neurons linking the buccal to the abdominal ganglia. At high-frequency firing rates during feeding behavior, the synaptic potentials may convert from excitatory to inhibitory [52,53].

*3.2. Avoidance and Approach*

Insect models [56] have been exploited to shed light on fundamental processes of memory formation and memory update in behavioral processes of attraction and aversion producing alternative choice responses [57,58]. Fruit flies can learn to associate an odor stimulus with a positive or negative consequence, such as food reward or electric shock punishment [59–61]. In the training phase, flies are typically exposed to two odors (differential conditioning) where one odor is perceived alone, whereas a second odor is presented together with either reward or punishment [59]. Once an association has formed between a stimulus and its consequence (reward, punishment), the learned anticipation of reward or punishment can be observed in a memory test that enforces a binary choice behavior (approach or avoidance) to positively and negatively reinforced stimuli [61]. A single learning trial [61,62] can be sufficient to form a stable memory. The so-called prediction error theory [57] describes a basic theoretical concept in the field of classical conditioning. It accounts for the fact that the efficacy of learning is determined by the discrepancy (or error) between the expected and the received reinforcement (reward or punishment). Re-exposing flies to a conditional stimulus (odor) after successful training in the absence of positive or negative reinforcement leads to a reduction in the previously learned behavior, a phenomenon called extinction learning [57,58,63–69], which is observed across invertebrate and vertebrate species. According to prediction error theory, extinction learning is driven by the repeated mismatch between the expected outcome (reward or punishment) of the initial learning (conditioning) phase and a different and unexpected new outcome, i.e., the absence of either positive or negative reinforcement, for example.

*3.3. Adaptation to the Unexpected*

In humans and in robots, extinction learning is of high relevance for behavioral adaptation to the unexpected. Such would include the ability to choose an alternative trajectory when the programmed one presents an unexpected obstacle, for example. Data from conditioning experiments suggest that two parallel but opposing memory traces coexist in the functional neural network architectures of biological reinforcement and extinction learning [70–76]. A minimalistic model of such circuitry has been proposed recently [74] to account for classical appetitive and aversive conditioning with memory extinction. This model is tailored to existing anatomical data, with two circuits of critical importance that exploit highly plastic synaptic connections between principal neurons (PN), functionally identified Kenyon cells (*K*), essential for olfactory learning and memory facilitated by dopamine-driven plasticity [72–75] of their signaling in response to odors, and functionally identified output neurons (*ON*) in separate and mutually inhibiting reward (attraction) and punishment (repulsion) learning pathways. Neuromodulation through recurrent network connections and the plasticity thereof permit implementing a simple mechanism that generates testable predictions in the temporal domain for the rapid encoding of associations of the conditioned stimulus with a reward or a punishment in single-trial learning (Figure 4). Each PN of the network model is activated at a random rate drawn from a uniform distribution within the range between 0 and 1. PNs are connected to *K* neurons via a first synaptic weight matrix (*W1*); each connection has a fixed synaptic weight. Activation of the *K* vector in the next layer results from the matrix product of the PN population vector and the respective weight matrix *W1*. *K* neurons are fully connected to the *ON* via a second weight matrix (*W2*). With all synaptic weights initially set to 0.01, the excitatory input to *ON4* and *ON1* mediating negative reinforcement results from a summation of inhibitory and excitatory input

$$ON4 = ON4+ + ON4- \tag{3}$$
$$ON1 = ON1+ + ON1- \tag{4}$$

whereas, for *ON3* and *ON2* mediating positive reinforcement, the activation rate is solely determined by excitatory input. KPOS and KNEG neurons receive excitatory feedback from *ON4* and *ON1* neurons, respectively. Reinforcing stimuli have an effect on both. A rewarding, unconditioned stimulus (positive reinforcement) generates excitatory input to the POS neurons while excitatory feedback from *ON1* to the NEG neurons is partially suppressed. Conversely, a punishing, unconditioned stimulus (negative reinforcement) generates excitation of NEG neurons, as well as the partial suppression of excitatory input from *ON4* to the POS neurons. The complete model and equations for all processing stages are given in [74].

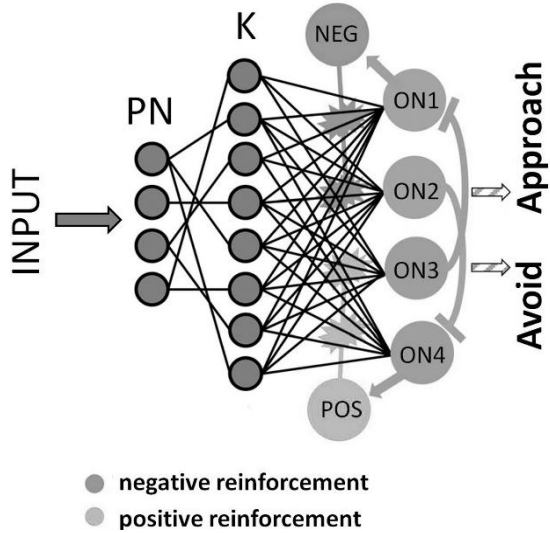

**Figure 4.** Invertebrate neural network model inspired by [74] for olfactory coding, olfactory memory formation during the negative or positive reinforcement of odors (reinforcement learning), and extinction learning when positive or negative reinforcement is no longer delivered. The neural network architecture and its functional properties account for transitions between approach and avoidance behaviors, and vice versa, in response to odors during extinction learning. The network consists of three fully connected functional layers of principal neurons (PN), reward (POS), and punishment (NEG) coding dopaminergic Kenyon neurons (K), and output neurons (ON), representing the three major stages of the olfactory pathway in *Drosophila*.

Similar neural network models based on a predictive form of Hebbian synaptic plasticity [75] account for a wide range of experiments on insect learning in uncertain environments including risk aversion. The predictive Hebbian model uses neuromodulatory influences to bias specific actions and to control synaptic plasticity. The neural substrates of prediction and reward [76] provide model accounts that have been in behavioral simulation [74–77].

### 4. Vertebrate Models of Learning for Cognitive Control

Evolution and individual brain development are open-ended processes of information increase and, as a consequence, information processing capacity [77], where an agent's capabilities of learning and acting, i.e., the level of agency, represent a functional compromise between stability and specificity, on the one hand, and the anticipation of external (environmental) change, on the other. Unsupervised reinforcement learning therein is a universal mechanism, widely used to explain behavior and behavioral control. It accounts for the lower-level adaptive learning in invertebrates illustrated in Section 3, and the higher-level learning for cognitive control in vertebrates including the nonhuman and human primate [78–89]. In vertebrate species, reward (reinforcement) learning consists of an agent learning specific values associated with specific states that constitute a so-called task state space [80–89]. The agent then uses the learnt knowledge to control the multiple-alternative choice of actions likely to lead to desired (reinforced) outcomes [79–82].

*4.1. Task State Learning and Control*

It has been proposed that neural networks in the mammalian orbitofrontal cortex [83–88] encode task states and task state spaces [82] during reinforcement learning. How the OFC acquires and stores this kind of information is not well understood. Neural network hypotheses and models [79–92] have attempted to propose and simulate cortical candidate mechanisms inspired by known functional properties of the primate brain. Neural network models based on reservoir computing represent a suitable approach for encoding task state information during reinforcement learning [79–82]. Reservoir networks [82] exhibit heterogeneous and dynamic activity patterns that can be exploited (Figure 5). Most reinforcement learning models account for human or animal behavior whilst assuming that the agent knows the task structure; yet, in the case of real agents (animals, humans, robots), the task structure needs to be learnt. It is critical for such a network to receive reward information as part of its input and, just as the orbitofrontal cortex receives converging sensory and reward inputs, the network is able to acquire task structure and support reinforcement learning by encoding combinations of sensory and reward events [81–83]. The network is based on the principles of a two-stage decision task where the agent (primate, robot) has to choose between two options A1 and A2. Their choices then lead to two intermediate outcomes B1 and B2 with different fixed probabilities. Choosing A1 is more likely to lead to B1, and choosing A2 is more likely to lead to B2. The final reward associated with a given choice is contingent only on the intermediate outcomes, and this contingency is reversed across trial blocks. Thus, the probability of getting a reward is higher for B1 in one trial block, and then becomes lower in the next while the probabilistic association between initial choices and intermediate outcomes never changes. The learning agent is not informed of the structure of the task and has to figure out the optimal choice response by tracking not only the final reward outcomes but also the intermediate outcomes. Further details, in the framework of a self-organizing neural network model for state encoding and reward association, can be found in [81]. The work described therein makes a compelling model case for the habitual process of reinforcement learning in interaction with specific goal-directed aspects by showing that such an interaction need not be coordinated by external arbitration. The principle of self-organization [81,82] plays an important part to such effect, as clarified later with regard to unsupervised control of robot and sensor learning.

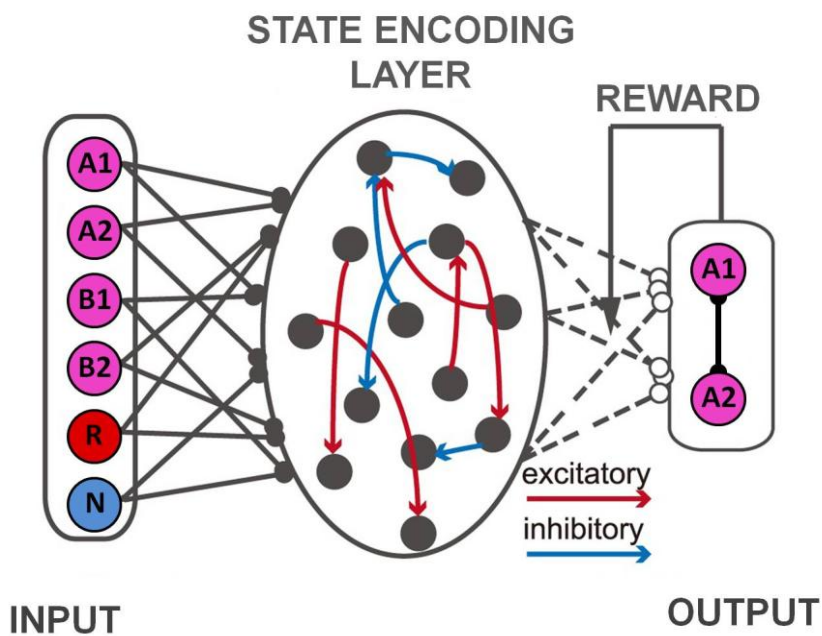

**Figure 5.** Cortical neural network model simulating reward (R) and non-reward associations with alternative choice decisions (A1, A2) leading to different consequences (B1, B2) that have equal probabilistic weight but change between trial blocks. A state-encoding layer between input and output simulates the excitatory and inhibitory neural mechanisms generating the conditional encoding (internal representation) of reward versus non-reward

contingencies associated with a given choice response (decision) during reinforcement learning.

Thus, biological learning algorithms have the possibility to resort to internal processing hierarchies in the formation of action sequences for habit learning and goal-directed actions, on the one hand, and the habituated sequencing of actions, on the other, through excitatory and inhibitory brain-inspired mechanisms [80–94]. This accounts for so-called event coding [81,84] during reinforcement learning in line with experimental findings [78–81].

*4.2. Memorizing Temporal Order*

Remembering the temporal order of a sequence of events is easy for humans and most mammals in everyday life. The underlying neuronal dynamics are self-organizing, as illustrated in models inspired by functional properties of the primate brain [2,3]. Overt human behavior and its full expression proceed on a timescale of seconds or minutes for longer tasks [95], which appears to contrast with the ultrafast millisecond timescale of neuronal processing in the primate brain. Sequence learning (Figure 6) in neural networks has been a model in terms of finely tuned temporal firing activities enabling the compression of slow behavioral sequences down to the millisecond timescale, which is that of synaptic plasticity. Mathematical analysis and computer simulations have produced the phenomenon of phase precession [95–99]. Within critically short synaptic learning windows, phase precession was found to improve temporal-order neural network learning [98,99]. Putative mechanisms for linking the millisecond timescale of synaptic plasticity to the slow timescale of behavior relate induction times of synaptic plasticity to spike-timing-dependent plasticity, a specific form of synaptic plasticity, taking into account the temporal order of presynaptic and postsynaptic spiking, on the one hand, and the slower firing rates of place cells [97–99], a specific class of location coding neurons, on the other. Such neurons start firing in specific patterns when an animal visits certain learned locations in its familiar environment. A 'learning window' constitutes the temporal intervals at which presynaptic and postsynaptic activities induce synaptic plasticity during learning, and model accounts have simulated precisely timed neural activity generated by phase precession, i.e., the successive across-cycle shift of from-late-to-early spike phases by comparison with a background oscillation [98]. Phase precession allows for a temporal compression of a sequence of behavioral events from the timescale of seconds to that of milliseconds [99–101], matching the widths of generic spike-timing-dependent plasticity (STDP) learning windows [102–104]. The processes of synaptic plasticity are described activity-dependent alterations of synaptic transmission efficiency (functional plasticity) resulting from or accompanied by changes in the structure and number of synaptic connections (structural plasticity). Information storage for memory representation is highly influenced by activity patterns of neurons and networks the timing of their firing activity. Both determine the plasticity potential of neurons by generating changes in their input-output characteristics. Excellent overviews of these synaptic mechanisms, from molecules to neural circuit integration, are provided in [98,99]. Precisely timed integration of spatial locations and the trajectories linking them has been accounted for by plastic mechanisms in the hippocampus, where overlapping place cell activities and their subsequent temporal compression determine the time windows for spike-timing-dependent synaptic plasticity (STDP). Detailed mathematical accounts for the putative synaptic learning rules in such models are provided in [98–102]. These examples illustrate how the network representation of action and event sequences is formed, modified, and modified again in time by the self-organizing mechanisms of synaptic learning and plasticity [103–105].

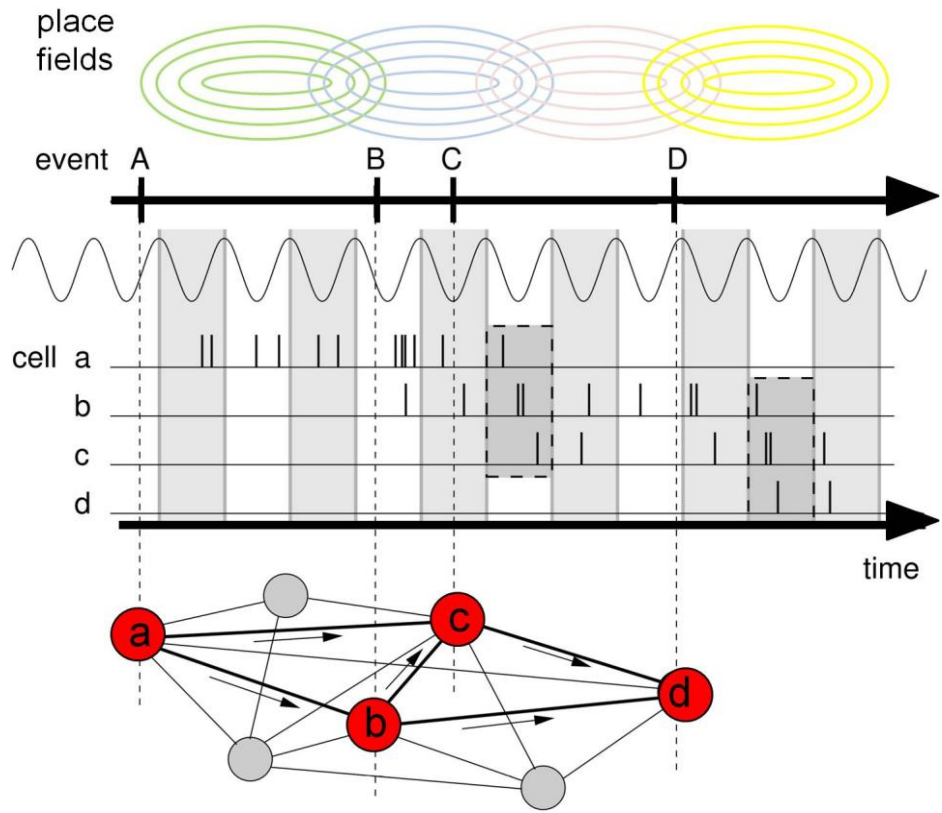

**Figure 6.** Encoding of spatial sequences in the mammalian brain through neural mechanisms of temporal compression, as described earlier [98,99]. The place fields (top), corresponding to oscillatory neural activities of place cells in the hippocampus, may spatially overlap, an important condition for network learning of trajectories.

*4.3. Self-Organization*

Biological learning is, as highlighted above using selected examples from invertebrates and vertebrates, by definition self-organizing. On this basis, the brain representation of agentic experience is generated by groups of highly interconnected neurons called cell assemblies in both invertebrate and vertebrate learning. In higher-order learning, allocation and storage of information in connected circuitry operate on the basis of synaptic weight adaptation in different types of synapses, networks, and functionally connected assemblies of networks. Synaptic plasticity is the basis of all cognition and memory [2,3]. While memory allocation is sometimes associated with the synaptic changes at feedforward synapses, storage with the adaptation of recurrent connections most likely involves both [103–105]. The functional principle through which memory allocation and storage is achieved, and the adaption of different synapses and networks involved is coordinated allows for reliable representation of multiple memories without disruptive interference between. This functional principle is that of self-organization [105–107]. As discussed in full detail elsewhere [106], there are seven key properties of self-organization in vertebrate brain systems: (1) modular connectivity, (2) unsupervised learning, (3) adaptive ability, (4) functional resiliency, (5) functional plasticity, (6) from-local-to-global functional organization, and (7) dynamic system growth. They are derived from insights in neurobiology, cognitive neuroscience, physics, and, in particular, Grossberg's [2] adaptive resonance theory (ART), which provides a mechanistic, mathematically supported, account of how self-organization achieves stability and functional plasticity while minimizing structural system complexity. The principle is exploited in Kohonen's [107] self-organizing map, a computationally parsimonious example of self-organizing, brain-inspired artificial neural network (ANN) recently employed in simulations of brain-like sensory learning for automatic (sensor or robot driven) detection of microscopic changes in physical environments [108–112]. The SOM has a functional architecture that formally corresponds to the nonlinear, ordered, smooth mapping of high-dimensional input data to representations in terms of a regular, low-dimensional array [107]. Any set of input variables can be defined as a real vector *x* of *n*-dimension. A parametric real vector $m_i$ of *n*-dimension is associated with each representation in the SOM, with the vector $m_i$ being a model, and the SOM an array of model representations. Assuming a general distance measure between *x* and $m_i$ given by $d(x,m_i)$, the map of an input vector *x* on the SOM array is then defined as the representation $m_c$ that best matches *x* yielding the smallest $d(x,m_i)$. During unsupervised learning, an input vector *x* is compared with all the $m_i$ to identify $m_c$. Euclidean distances $\|x–m_i\|$ define $m_c$. Models topographically close in the map, up to a certain geometric distance, indicated by $h_{ci}$, activate each other to learn from their joint input *x*. This results in a local relaxation or smoothing effect on the models in the neighborhood and leads to global ordering. Self-organized (SOM) learning is inspired by the Hebbian principles summarized in Equations (1) and (2). SOM learning can be expressed in the form

$$m(t+1) = m_i(t) + \alpha(t)h_{ci}(t)[x(t) - m_i(t)], \qquad (5)$$

where $t = 1, 2, 3 ...$ represents an integer, the discrete-time coordinate $h_{ci}(t)$ is the neighborhood function, a smoothing kernel defined across the ma which converges toward zero with time, and $\alpha(t)$ is the learning rate, which also converges toward zero with time. This particular form of unsupervised learning uses the *winner-take-all* principle, where each image input vector *x* is matched to its best matching model within the map $m_c$. Similarly [105], recent network simulations and phase space analyses have revealed that the interplay between long-term synaptic plasticity and homeostatic synaptic scaling simultaneously self-organizes the adaptation of feedforward and recurrent synapses such that a new stimulus forms a new memory wherein different stimuli are assigned to distinct cell assemblies. The resulting dynamics can reproduce experimental in vivo data relative to neuronal excitability and network connectivity, as well as their influence on memory formation. Thus, it is made clear that the few fundamental Hebbian synaptic mechanisms follow self-organizing principles for plastic and, at the same time, stable representation in biological neural circuitry.

*4.4. Toward "Intelligent" Robotics*
The Hebbian learning principles are a rich source inspiration for the design of biologically plausible lower- and higher-level, multifunctional control in robotics. On the basis of functionally identified neurons and connectivity principles in combination with biomechanical parameters driving multifunctional behavior, testable experimental hypotheses are generated, which then in return clarify the biological mechanisms and purposes of multifunctionality. A biologically relevant control framework is likely to be computationally efficient in the direct, real-time control of artificial robotic systems. At the same time, these systems can provide deeper functional insights into the biological system that serves as the model, building a bridge between systems neuroscience and robotics. The Hebbian synaptic learning model and its implementations account for neuromodulatory effects in invertebrates and vertebrates, as shown above using examples. Elementary (reflex-like) learning in invertebrates can be directly exploited for the control of robot motor learning in the absence of reward principles that account for motivational representation. As shown above using examples from invertebrates and vertebrates, the control of high-level learning for precisely timed and motivated movements and actions, including the avoidance of obstacles and choice of pertinent alternatives in response to the unexpected, relies on synaptic plasticity and the neural substrates of reinforcement learning through punishment or reward. The functional principle of self-organization offers computational solutions for unsupervised learning algorithms toward autonomous robot function and control. One of the advantages of the biological models reviewed above is that they all can be tested and have been in behavioral simulations. Not all their aspects are currently exploited in robotics for developing new functional architectures. How this may become possible in the near future is illustrated in the next section by discussing examples of current developments in "intelligent" robotics.

## 5. Current Developments in Brain-Inspired Robot Control

Multifunctional control in real time is a critical target in intelligent robotics. Combined with behavioral flexibility, such control enables real-time robot navigation and adaption to complex, often changing environments. Multifunctionality is observed across a wide range of living species and behaviors. As made clear above, even seemingly simple organisms such as invertebrates demonstrate multifunctional control. Living systems rely on the ability to shift from one behavior to another, and to vary a specific behavior for successful action under changing environmental conditions. Truly multifunctional control is a major challenge in robotics. A plausible approach is to develop a methodology that maps multifunctional biological system properties onto simulations [113] to potentiate rapid prototyping and real-time simulation of solutions (control architectures). The resulting controllers can then be tested and improved accordingly by comparison with the original biological system. Their relative effectiveness as simulated controllers of an artificial device (robot) is then evaluated on the basis of clear criteria (benchmarks). Below, some examples of current development in this direction, inspired by biological learning mechanisms discussed in the previous sections, are highlighted.

*5.1. Repetitive or Rhythmic Behavior*

Hybrid model frameworks combining synaptic plasticity-dependent neural firing with simple biomechanics at speeds faster than real time illustrate how invertebrate learning directly inspires "intelligent" robotics [114,115]. Such frameworks exploit a multifunctional model of *Aplysia* feeding rhythms, which are capable of repeatedly reproducing three types of behavior: biting, swallowing, and rejecting. These simulate behavioral switching in response to external sensory cues. Model approaches incorporate synaptic learning and neural connectivity in a simple mechanical model of the feeding apparatus [116], with testable hypotheses in the context of robot movement control. As explained in detail in Section 3.1, the neural networks that govern feeding in *Aplysia* include motor neurons and cerebral–buccal target interneurons. Learning-induced synaptic plasticity in such modular circuitry controls behavioral switching (Figure 7), as recently simulated in biologically inspired model approaches directly exploitable for multifunctional robot control. For the model equations, the reader is referred to [116–120]. This modeling framework can be extended to a variety of scenarios for multifunctional robot movement and rhythm control, and it has several advantages. It allows rapid simulation of multifunctional behavior and it includes the known functional circuitry and simplified biomechanics of peripheral anatomy. The direct relationship with the underlying neural circuitry makes it possible to both generate and test specific neurobiological hypotheses. The relative simplicity of the network (Figure 7) makes it attractive as a basis for robot control. Unlike other artificial neural network architectures, synthetic nervous systems are explainable in terms of structures directly informing the functional system output [121–125]. Although the connections and trained weights of other artificial neural networks may provide similar control capabilities, these, unlike synthetic nervous systems, have to be trained on large datasets. The very strength of synthetic nervous systems is that they use a restricted, functionally identified set of biological neuron dynamics, thereby generating robust control without the need for additional training [125]. Neural network learning-inspired robotics include reactive systems emulating reflexes, neural oscillators to generate movement patterns, and neural networks for filtering high-dimensional sensory information [126]. To such effect, biologically motivated neural-network-based robot controllers, inspired by control structures in the sensory brain, where information is routed through the network using facilitating dynamic synapses with short-term plasticity, have been proposed [123–128]. Learning occurs through long-term synaptic plasticity using temporal difference learning rules to enable the robot to learn to associate a given movement with the correct, i.e., appropriate as defined, input conditions. Self-organizing network dynamics [127,129] provide memory representations of the environments that the robot encounters.

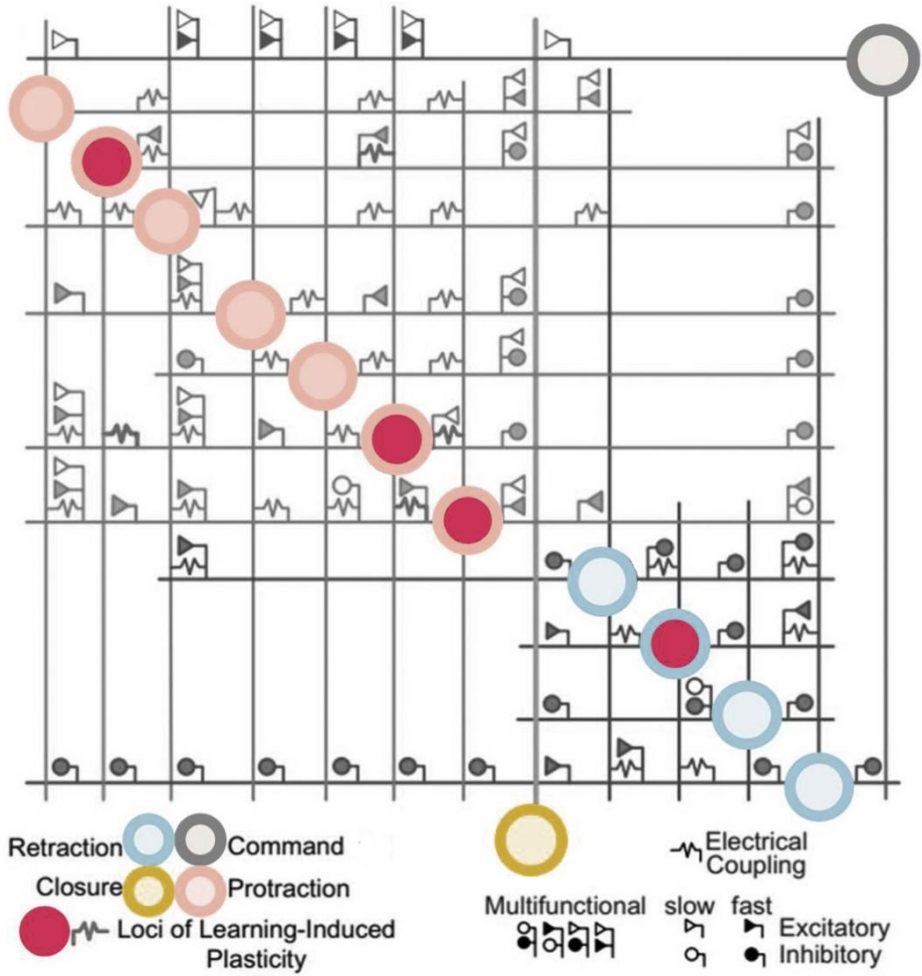

**Figure 7.** Model circuitry, adapted from [116,119,120], for multifunctional robotic movement command/control based on the functional neuroanatomy and synaptic plasticity of *Aplysia* motor and interneurons (see Figure 3).

*5.2. Sensorimotor Integration*

Recent progress in neuromorphic sensory systems which mimic the biological receptor functions and sensorial processing [129–132] trends toward sensors and sensory systems that communicate through asynchronous digital signals analogous to neural spikes [127], improving the performance of sensors and suggesting novel sensory processing principles that exploit spike timing [128], leading to experiments in robotics and human–robot interaction that can impact how we think the brain processes sensory information. Sensory memory is formed at the earliest stages of neural processing (Figure 8), underlying perception and interaction of an agent with the environment. Sensory memory is based on the same plasticity principles as all true learning, and it is, therefore, an important source of intelligence in a more general sense. Sensory memory is consolidated while perceiving and interacting with the environment, and a primary source of intelligence in all living species. Transferring such biological concepts into electronic implementation aims at achieving perceptual intelligence, which would profoundly advance a broad spectrum of applications, such as prosthetics, robotics, and cyborg systems [129]. Moreover, transferring biologically intelligent sensory processing into electronic implementations [130–132] achieves new forms of perceptual intelligence (Figure 8). These have the potential to profoundly advance a broader spectrum of applications in robotics, artificial intelligence, and control systems.

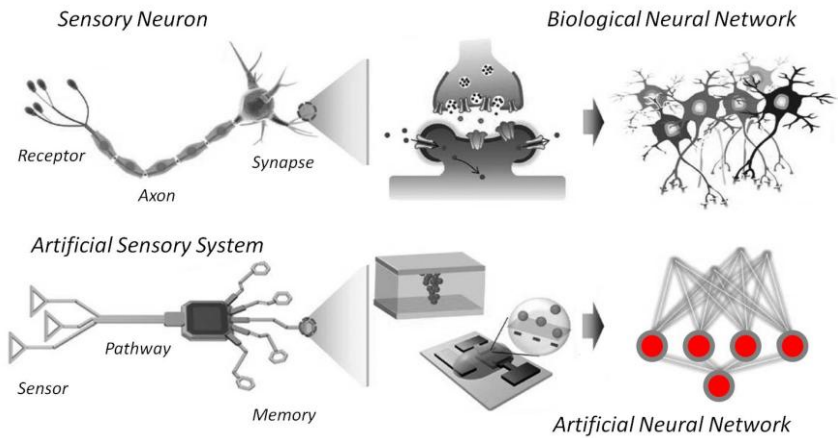

**Figure 8.** Biomimetic sensory systems: from biological synapses to artificial neural networks for novel forms of "perceptual intelligence".

These new, bioinspired systems offer unprecedented opportunities for hardware architecture solutions coupled with artificial intelligence, with the potential of extending the capabilities of current digital systems to psychological attributes such as sensations and emotions. Challenges to be met in this field concern integration levels, energy efficiency, and functionality to shed light on the translational potential of such implementations. Neuronal activity and the development of functionally specific neural networks in the brain are continuously modulated by sensory signals. The somatosensory cortical network [133] in the primate brain refers to a neocortical area that responds primarily to tactile stimulation of skin or hair. This cortical area is conceptualized in the current state of the art [133–136] as containing a single map of the receptor periphery, connected to a cortical neural network with modular functional architecture and connectivity binding functionally distinct neuronal subpopulations from other cortical areas into motor circuit modules at several hierarchical levels [133–136]. These functional modules display a hierarchy of interleaved circuits connecting, via interneurons in the spinal cord, to visual and auditory sensory areas, and to the motor cortex, with feedback loops and bilateral communication with the supraspinal centers [135–137]. This enables 'from-local-to-global' functional organization [134], a ground condition for self-organization [106,107], with plastic connectivity patterns that are correlated with specific behavioral variations such as variations in motor output or grip force, which fulfills an important adaptive role and ensures that humans are able to reliably grasp and manipulate objects in the physical world under constantly changing conditions in their immediate sensory environment. Neuroscience-inspired biosensor technology for the development of robot-assisted minimally invasive surgical training [138–143] is a currently relevant field of application here as it has direct clinical, ergonomic, and functional implications, with clearly identified advantages over traditional surgical procedures [144,145]. Individual grip force profiling using wireless wearable (gloves or glove-like assemblies) sensor systems for the monitoring of task skill parameters [138–141] and their evolution in real time on robotic surgery platforms [141–143,146–148] permits studying the learning curves [140–142] of experienced robotic surgeons, surgeons with experience as robotic platform tableside assistants, surgeons with laparoscopic experience, surgeons without laparoscopic experience, and complete novices. Grip force monitoring in robotic surgery [146–148] is a highly useful means of tracking the evolution of the surgeon's individual force profile during task execution. Multimodal feedback systems may represent a slight advantage over the not very effective traditional feedback solutions, and the monitoring of individual grip forces of a surgeon or a trainee in robotic task execution through wearable multisensory systems is by far the superior solution, as real-time grip force profiling by such wearable systems can directly help prevent incidents [146,147] because it includes the possibility of sending a signal (sound or light) to the surgeon before their grip force exceeds a critical limit, and damage occurs. Proficiency, or expertise, in the control of a robotic system for minimally invasive surgery is reflected by a lesser grip force during task execution, as well as by a shorter task execution times [146–148]. Grip forces self-organize progressively in a way that is similar to the self-organization of neural oscillations during task learning, and, in surgical human–robot interaction, a self-organizing neural network model was found to reliably account for grip force expertise [149].

*5.3. Movement Planning*

To move neural processing models for robotics beyond reactive behavior, the capacity to selectively filter relevant sensory input and to autonomously generate sequences of processing steps is critical, as in cases where a robot has to search for specific visual objects in the environment, and then reach for these objects in a specific, instructed serial order [150,151]. In robotic tasks where the simultaneous control of object dynamics and internal forces exerted by the robot limb(s) to follow a trajectory with the object attached to it is required, plasticity and adaptation permit to deal with external perturbations acting on the robot–object system. On the basis of mere feedback through the internal dynamics of an object, a robot is, like a human, able to relate to specific objects with a very specific sensorimotor pattern. When the object-specific dynamical patterns are combined with hand coordinates obtained from a camera, dedicated hand-eye coordination self-organizes spontaneously [152–154] without any higher-order cognitive control. Robots are currently not capable of any form of genuine cognition. Cognition controls behavior in living brains, where sensing and acting are no longer linked directly to ensure control, as is the case for any robot currently, including humanoids. When an action is based on sensory information that is no longer directly available in the processing loop at the time where action is to ensue, the relevant information must be represented in a memory structure, as it is in any living brain. Information for the control of action then becomes abstracted from sensor data through the neural memory representations and mechanisms of memory-based decision making [150]. Plastic mechanisms in neural network-based control architectures (Figure 9) effectively contribute to the learning of dynamics of robot–object systems, enabling adaptive corrections and/or offset detection.

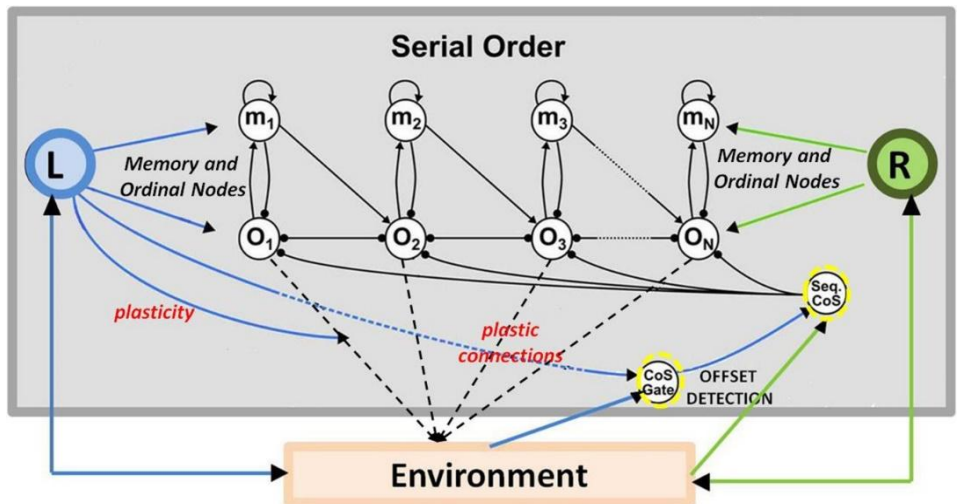

**Figure 9.** Biological learning-inspired processing model inspired by [150] for robotic control beyond reactive behavior, with a capacity to selectively filter relevant sensory input and to autonomously generate sequences of processing steps. The illustration here shows the dynamic neural network architecture for the control of perceptual learning (L), memory storage of objects/actions, their serial order, and recall (R) through node structures with plasticity enabled connections. Selectively gated feedback is enabled through computational nodes (outlined in yellow here) for the updating of sensory representations (such as offset detection) as a function of changes in input from the environment.

This allows for progressive error reduction by incorporating distributed synaptic plasticity according to feedback from actual movements in the given environment. It has been shown previously that such feedback processes are omnipresent in voluntary motor actions of human agents [154], where rapid corrective responses occur even for very small disturbances that approach the natural variability of limb movements. Robot control toward autonomy [155] ultimately implies that the robot generalizes across time and space, is capable of stopping when an element is missing, and updates a planned action sequence autonomously in real time when a scenario suddenly changes. Using biologically plausible neural learning, the flow of behavior generated can emerge new neural system dynamics through self-organization without any further control or supervision algorithm(s). In robotic control based on biologically inspired neural network learning, the universal training method is based on Hebbian synaptic learning. Several variants of the latter are discussed and compared, with detailed equations, in [150]. The neural network dynamics described therein can, in principle, be combined with other network structures that receive reward information as part of their input in an extended model approach based on known functional dynamics of the mammalian brain. As discussed in previous sections, the orbitofrontal cortex receives converging sensory and reward inputs, which makes the network as a whole capable of acquiring task structure and support reinforcement learning by encoding combinations of sensory and reward events [81,82]. Such networks possess self-organizing state-encoding dynamics of the type shown here above (Figure 5), based on the principle of multiple-stage decision tasks, where a human agent or robot has to choose between decisions (options) and their consequences. More knowledge from and interaction between the fields of cognitive neuroscience and robotics are needed here to further explore existing possibilities.

## 6. Conclusions

Living organisms have a long evolutionary history of structure and function ensuring their survival in natural environments. Such adaptation relies on biological adaptive learning, from single synapse to networks, which is unsupervised and self-organizing. Table 1 gives a schematic overview of selected references, highlighted here as a function of the type of agency (behavior or function), species exploited (model), and level of self-organizing functional complexity for the development of self-organizing neural network solutions (control) toward "intelligent" robotics. By emulating biological mechanisms for the development of electronic systems for sensorimotor control, the approaches reviewed here push humanoid robots, exoskeletons, and similar electronic systems toward increasing levels of autonomy. Artificial synapses have emulate the brain's plasticity with much simpler, less costly structures than most other traditional computing methods. They, therefore, offer promising perspectives for future robotic and neuromorphic systems. The working mechanisms of biological synapses and brain plasticity demonstrate the learning and memory potential of extremely simple and highly complex functions in living organisms. Some of them already comprise the sensory systems of robots. Synaptic learning can be used to control artificial nerves and muscles that have the same working mechanism as biological ones, and new models derived from brain learning can breathe lifelike motion into mobile robots. In the near future, neuromorphic systems are expected to become vital components of robots and electronic applications, including biocompatible neural prosthetics, exoskeletons, soft humanoids, and integrated cybernetics, exploiting natural sensory and memory systems to project robotics into the future.